\documentclass[conference]{IEEEtran}
\IEEEoverridecommandlockouts
\usepackage{cite}
\usepackage{amsmath,amssymb,amsfonts}
\usepackage{algorithmic}
\usepackage{graphicx}
\usepackage{textcomp}
\usepackage{xcolor}
\usepackage{hyperref}
\usepackage{amsthm}
\usepackage{color}
\usepackage{textcomp}
\usepackage{subfig}
\usepackage[T1]{fontenc}

\pdfoutput=1

\theoremstyle{plain}

\theoremstyle{definition}
\newtheorem{defn}{Definition} 
\newtheorem{prop}{Property} 

\def\BibTeX{{\rm B\kern-.05em{\sc i\kern-.025em b}\kern-.08em
    T\kern-.1667em\lower.7ex\hbox{E}\kern-.125emX}}
\begin{document}

\title{The Colliding Reciprocal Dance Problem:\\A Mitigation Strategy with Application to Automotive Active Safety Systems$^*$\thanks{$^*$Extended abstract for the 2019 Northeast Robotics Colloquium.}}

\author{\IEEEauthorblockN{Jeffrey Kane Johnson}
\IEEEauthorblockA{\textit{Maeve Automation} \\
Pittsburgh, USA\\
\href{mailto:jeff@maeveautomation.com}{jeff@maeveautomation.com}}
}

\maketitle



\section{Introduction}

Anyone who has walked through crowded areas is likely familiar with a phenomenon where two people become unable to pass each other even if sufficient space for passage exists. This happens when a lack of coordination results in attempts to take mutually incompatible actions. The repeated selection of incompatible actions results in a sort of oscillating deadlock that we refer to as a ``Reciprocal Dance."

For pedestrians, a reciprocal dance is typically just a nuisance. 
However, for agents with inertial constraints a reciprocal dance can lead to collision, with severe, or even fatal, consequences. This abstract presents this colliding variant of the reciprocal dance, how it arises, a mitigation strategy, and an example in the context of automotive active safety.

\section{Formalizing and Mitigating the Colliding Reciprocal Dance Problem}


\subsection{Selective Determinism}

In many multi-agent navigation problems, it can be assumed that agents will prefer self preservation to goal-directed progress. This assumption can be exploited to factor behavioral interaction effects out of sequential decision making to make an otherwise intractable problem tractable. {\em Selective Determinism}~\cite{phd} is a framework that formalizes this factorization using a property called {\em Stopping Path Disjointness}, which is defined in terms of a {\em Stopping Path} and {\em Stopping Region}:

\begin{figure}%
    \centering
    \subfloat[Bird's eye view]{{\includegraphics[width=0.21\textwidth]{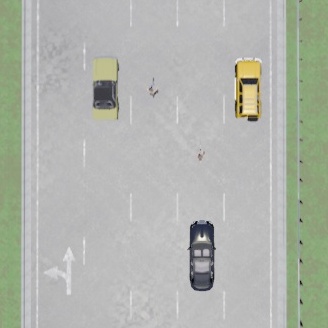} }}%
    \qquad
    \subfloat[Front camera view]{{\includegraphics[width=0.21\textwidth]{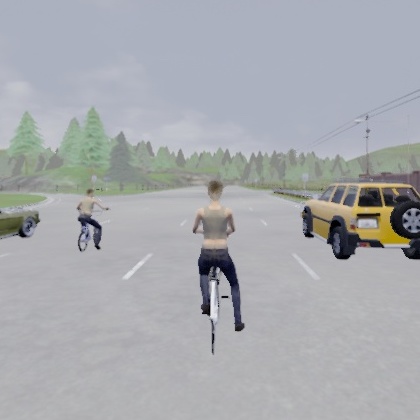} }}%
    \caption{The vehicle approaches a stopped cyclist, relying on the collision avoidance system to modulate control.}%
    \label{fig:scenario}%
\end{figure}

\begin{defn}\label{def:stopping-path}
A {\em Stopping Path} is the minimum space an agent needs to occupy in order to come to a stop\footnote{In this context ``stop" could mean zero relative, or zero absolute, velocity. For further discussion, see~\cite{phd}.} along a given path. Also referred to as a {\em contingency plan}.
\end{defn}

\begin{defn}\label{def:stopping-region}
A {\em Stopping Region} is the disjoint union of all stopping paths defined over the set of all followable paths.
\end{defn}

\begin{prop}\label{prop:sp-disjointness}
{\em Stopping Path Disjointness} is the property that each agent has least one stopping path disjoint from the stopping regions of all other agents.
\end{prop}


Property~\ref{prop:sp-disjointness} guarantees for all agents the existence of a collision-free contingency plan. Selective determinism provides a mechanism for all agents to maintain the property independently but requires that they invoke a contingency if not doing so will violate it. It is this, rather than policy optimality, that ensures safety, allowing agents to trade optimal interaction models for computationally simpler ones. However, this trade-off also allows certain pathological situations.

\subsection{Reciprocal Dance as Pathological Selective Determinism}

In many applications, interactions tend to be simple, and incompatibilities transient, so agents with simplified interaction models generally navigate successfully. But what if interaction models {\em consistently} interact incorrectly? This pathological interaction case can provide a slightly more formal definition of a reciprocal dance:

\begin{defn}\label{def:rd}
A {\em Reciprocal Dance} is a situation in selective determinism when mutually incompatible interaction models cause a deadlock of repeated contingency invocation.
\end{defn}

\subsection{The Colliding Reciprocal Dance}

A reciprocal dance can be {\em colliding} if it occurs when Property~\ref{prop:sp-disjointness} is not maintained. 
Unfortunately, maintaining this property can be non-trivial. 
In order for it to hold, the boundary of a {\em computed} stopping region must be conservative with respect to the boundary of the {\em true} stopping region. 
Inertial constraints can make this difficult to guarantee because the resulting stopping regions tend to sweep out large, complex volumes for which it can be difficult to approximate usefully conservative bounds. 
Fortunately, a simple strategy can help mitigate the effects of non-conservative approximations.


\subsection{A Mitigation Strategy: Constraint Tightening}

As a mitigation strategy, we bias an agent's dynamic constraints to those of its current stopping region. From the definition of a stopping region, this is conceptually straightforward: The agent has some set of paths available to it, and, for each path and current agent state, it has {\em nominal} dynamic constraints that allow the path to be followed. The stopping region has the same set of paths coupled with {\em contingency} dynamic constraints that most quickly bring the agent to a stop. {\em Constraint tightening}, our mitigation strategy, is the process of adaptively adjusting the bounds of the nominal constraints toward those of the contingency constraints with respect to some notion of proximity. This strategy is effectively a type of adaptive damping that can reduce the risk of collisions without significantly limiting non-emergency behavior.


%


\section{Automotive Active Safety Systems}

Selective determinism decouples of the navigation task into independent collision avoidance and guidance tasks. In an automobile, the collision avoidance task could be implemented as an active safety system and the guidance task assigned to a human driver. Because of vehicle inertial constraints, such a system would be at high risk for colliding reciprocal dances and thus is ideal for demonstrating our mitigation strategy.


We use the CARLA simulator~\cite{carla} to implement a collision avoidance system on a human-guided vehicle. We conduct trials that have the human command the vehicle at full throttle into a stationary cyclist\footnote{This emulates a driver asleep at the wheel or distracted by a cell phone.} as shown in Figure~\ref{fig:scenario}. In the simulator, vehicle dynamics are computed using PhysX~\cite{physx}. For collision avoidance, we approximate vehicle motion with a simpler constant acceleration model\footnote{For simplicity, we restrict vehicle motion here to only straight line. A demonstration using the joint steering and throttle space is referenced in \S\ref{sec:conclusion}.}. We set the model deceleration to 90\% of peak deceleration so that the vehicle will typically slow more quickly than the system predicts. While this should result in conservative behavior, we nevertheless expect that the discrepancy in vehicle models will result in colliding reciprocal dance situations, even for this simple scenario.

We compare three mitigation strategies:

\begin{enumerate}
\item \textbf{Constraint Tightening:} Nominal constraints are proximity-biased to contingency constraints.\label{mit:rsr}
\item \textbf{Conservative Deceleration:} The simplified motion model minimum is set to 80\% of peak achievable.\label{mit:cons}
\item \textbf{None:} The simplified motion model is used as-is.\label{mit:none}
\end{enumerate}

\begin{figure}[htbp]
\centerline{\includegraphics[width=0.5\textwidth]{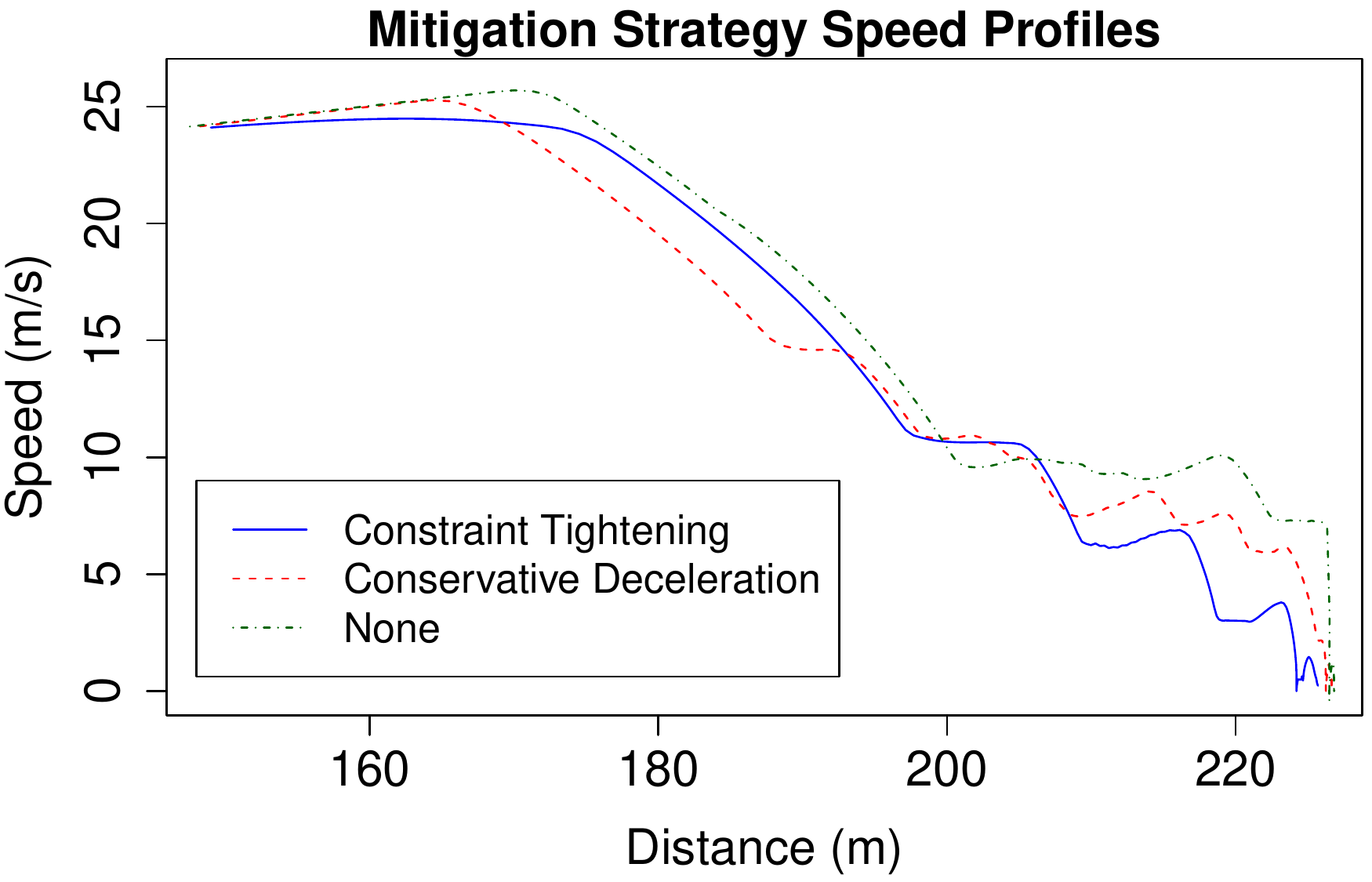}}
\caption{Vehicle speed as a function of path position. The cyclist is located at approximately position $225m$. 
}
\label{fig:speed-profiles}
\end{figure}

Speed profile plots of the vehicle for each strategy are shown in Figure~\ref{fig:speed-profiles}. Note the oscillating speeds toward the ends of the plots. This is due to the guidance signal disregarding agent interaction and the collision avoidance system repeatedly invoking contingencies. Classic reciprocal dance behavior.

Strategy~\ref{mit:rsr} permits the vehicle to maintain a higher speed along a greater extent of the path than Strategy~\ref{mit:cons} until approximately position $200m$, when it becomes more conservative. As the vehicle nears the cyclist, Strategy ~\ref{mit:rsr} keeps the vehicle slow and safely distant while Strategy~\ref{mit:cons} maintains relatively high speed until close proximity, only then invoking significant deceleration. This demonstrates the utility of the proposed strategy: the adaptivity permits the vehicle greater dynamic range when it is safe, and more conservative dynamic range when needed.
Finally, note that Strategy~\ref{mit:none} exhibits a sharp speed decline at the end of the plot. This is caused by collision with the cyclist. In the absence of a mitigation strategy, the simple motion model resulted in a colliding reciprocal dance.

\section{Conclusion}\label{sec:conclusion}

The colliding reciprocal dance problem is a common problem in mobile robotics systems, especially those with strong inertial constraints. By formulating the problem in a principled way, we have devised a principled approach to mitigating it. The proposed mitigation strategy is beneficial over more na\"{i}ve approaches because it provides adaptive behavior in order to maintain both safety and flexibility.

Data sets for the trials are available in ROS\cite{ros} bag format, along with RViz config for viewing, here: \href{https://maeveautomation.com/data-sets/\#e02}{https://maeveautomation.com/data-sets/\#e02}. A video of the system operating in the full steering and throttle space is available here: \href{https://youtu.be/DD4NfcHxpmY}{https://youtu.be/DD4NfcHxpmY}

\bibliographystyle{IEEEtran}
\bibliography{refs}

\end{document}